\documentclass[sigconf,nonacm]{acmart}

\AtBeginDocument{%
  \providecommand\BibTeX{{%
    \normalfont B\kern-0.5em{\scshape i\kern-0.25em b}\kern-0.8em\TeX}}}

\copyrightyear{2025}
\acmYear{2025}
\setcopyright{acmlicensed}\acmConference[WSDM '25]{Proceedings of the Eighteenth ACM International Conference on Web Search and Data Mining}{March 10--14, 2025}{Hannover, Germany}
\acmBooktitle{Proceedings of the Eighteenth ACM International Conference on Web Search and Data Mining (WSDM '25), March 10--14, 2025, Hannover, Germany}
\acmDOI{10.1145/3701551.3703497}
\acmISBN{979-8-4007-1329-3/25/03}

\usepackage{algorithm}
\usepackage[noend]{algpseudocode}
\usepackage{orcidlink}
\usepackage{dsfont}
\usepackage{bbold}

\newtheorem{remark}{Remark}
\newcommand{\candidatesset}{\mathcal{C}}

\newcommand\restr[2]{{
  \left.\kern-\nulldelimiterspace 
  #1 
  \littletaller 
  \right|_{#2} 
  }}
\newcommand{\littletaller}{\mathchoice{\vphantom{\big|}}{}{}{}}
\newcommand{\besttext}{\text{\textit{best}}}
\newcommand{\goodtext}{\text{\textit{good}}}
\newcommand{\addtext}{\text{\textit{add}}}
\DeclareMathOperator*{\argmax}{arg\,max}

\newcommand{\ci}{\mathrel{{\scalebox{1.07}{$\perp\mkern-10mu\perp$}}}}

\begin{document}

\title{The Initial Screening Order Problem}

\author{Jose M. Alvarez}
\orcid{0000-0001-9412-9013}
\affiliation{
  \institution{KU Leuven}
  \city{Leuven}
  \country{Belgium}
}
\email{josemanuel.alvarez@kuleuven.be}
\authornote{This work was done while being a Research Fellow at the University of Pisa.}

\author{Antonio Mastropietro}
\orcid{0000-0002-8823-0163}
\affiliation{
  \institution{University of Pisa}
  \city{Pisa}
  \country{Italy}
}
\email{antonio.mastropietro@di.unipi.it}

\author{Salvatore Ruggieri}
\orcid{0000-0002-1917-6087}
\affiliation{
  \institution{University of Pisa}
  \city{Pisa}
  \country{Italy}
}
\email{salvatore.ruggieri@unipi.it}

\renewcommand{\shortauthors}{Jose M. Alvarez, Antonio Mastropietro, and Salvatore Ruggieri}

\begin{abstract}
    We investigate the role of the initial screening order (ISO) in candidate screening. The ISO refers to the order in which the screener searches the candidate pool when selecting $k$ candidates. Today, it is common for the ISO to be the product of an information access system, such as an online platform or a database query. The ISO has been largely overlooked in the literature, despite its impact on the optimality and fairness of the selected $k$ candidates, especially under a human screener. We define two problem formulations describing the search behavior of the screener given an ISO: the best-$k$, where it selects the top $k$ candidates; and the good-$k$, where it selects the first good-enough $k$ candidates. To study the impact of the ISO, we introduce a human-like screener and compare it to its algorithmic counterpart, where the human-like screener is conceived to be inconsistent over time. Our analysis, in particular, shows that the ISO, under a human-like screener solving for the good-$k$ problem, hinders individual fairness despite meeting group fairness, and hampers the optimality of the selected $k$ candidates. This is due to position bias, where a candidate's evaluation is affected by its position within the ISO. We report extensive simulated experiments exploring the parameters of the best-$k$ and good-$k$ problems for both screeners. Our simulation framework is flexible enough to account for multiple candidate screening tasks, being an alternative to running real-world procedures. 
\end{abstract}




\maketitle

\section{Introduction}
\label{sec:Introduction}

Candidate screening is a complex, human-dependent task. 
It consists of a decision-maker or user, which we refer to as the \textit{screener}, tasked with selecting $k$ candidates from a candidate pool. 
Common candidate screening processes include the evaluation of resumes for a job interview \cite{Pisanelli2022_YourCV} or application packages for college admission \cite{SukumarMH18_PeacanPie}. 
The screener usually evaluates the candidate pool using limited information and under strict time constraints.
Information access systems (IAS), such as online platforms like LinkedIn and database queries like Taleo, play a central role today in candidate screening by allowing screeners to search more efficiently the candidate pool.
Enabled by machine learning (ML) \cite{DBLP:journals/corr/abs-2307-03195}, IAS often present candidates according to an estimated relevance or, at least, according to a relevant characteristic chosen by the screener \cite{DBLP:books/aw/Baeza-YatesR99}.
An industry around algorithmic candidate screening has emerged in recent years \cite{DBLP:journals/air/WillKL23}, though poor fairness results \cite{DBLP:conf/fat/WilsonG0MBSTP21,Wehner20,Sonderling22,Raghavan2020AlgortihmicHiring,DBLP:journals/datamine/RheaMDSSSKS22,jintelligence9030046,DBLP:journals/patterns/SloaneMC22, Fabris2023_DBLP:journals/corr/abs-2309-13933} and calls to consider the behavior of the human user \cite{Carricco2018EUHumanCentred, DBLP:conf/aaai/Ruggieri0PST23, bringas2022fairness, DBLP:journals/ethicsit/AlvarezCEFFFGMPLRSSZR24} continue to drive research on the social impact of IAS.

In this paper, we investigate the role of the \textit{initial screening order} (ISO) on the set selection problem implicit in the task of candidate screening. 
The ISO refers to the order in which the candidates appear in the candidate pool.
It can be chosen by or provided to the screener enabled via an IAS \cite{DBLP:journals/ftir/Ekstrand0B022}.
We develop a utility-based framework to understand the search behavior of the screener when going over the ISO, and use it to implement extensive simulations that study the influence of the ISO.
We find that the ISO can impact the optimality (i.e., choosing the best candidates) and fairness (i.e., treating similar candidates similarly) of the selected $k$ candidates, especially when the screener is human.
This is mainly because of the \textit{position bias} inherent to the ISO.
Here, position bias refers to the penalty (or premium) a candidate experiences due to where it falls on the ISO, as humans are predisposed to favor the items placed at the top of a list \cite{DBLP:journals/cacm/Baeza-Yates18, 10.1093/qje/qjr028}.

We motivate the ISO problem further in Section \ref{sec:Generali} based on our collaboration with an European company.
In Section~\ref{sec:ProblemFormulation}, we describe how the screener searches the ISO for an optimal and fair set of $k$ candidates w.r.t. a representational quota $q$ of protected candidates.
We define two problem formulations. 
In the best-$k$ the screener selects the $k$ top candidates, whereas in the good-$k$ the screener selects the $k$ first good-enough candidates fitting some minimum candidate quality measure.
We devise algorithmic solutions for both problems.
The good-$k$ is noteworthy as it allows the screener to partially search the candidate pool; the set selection problem formulation often assumes a full search.
In Section~\ref{sec:HumanScreener}, we analyze the algorithmic and human-like screeners to understand the impact of the ISO when a human is involved. 
The former refers to a consistent screener; the latter refers to an inconsistent screener whose evaluation of candidates suffers over time due to the fatigue of performing a repetitive task.
In Section~\ref{sec:Experiments}, we enhance our analysis of these two screeners through simulations that mimic multiple screening settings.
Our results confirm the role of position bias inherent to the ISO and raise new fairness concerns.
For instance, we find that the human-like screener violates individual fairness by not evaluating similar candidates similarly \cite{DBLP:conf/innovations/DworkHPRZ12} while still meeting the quota $q$, and that the algorithmic screener misses the best candidate depending on its search procedure.
In Section~\ref{sec:Discussion}, we conclude by discussing the limitations and extensions of our work.\footnote{We provide additional material in the Appendix.}

Our work is the first to formalize the ISO problem.
Our main contributions are threefold.
\textit{First}, we formalize the role of the ISO in two search behaviors of the screener with the best-$k$ and good-$k$ problems.
\textit{Second}, we introduce a human-like screener and compare it theoretically and experimentally to its algorithmic counterpart.
\textit{Third}, we provide a flexible simulation tool for studying the ISO problem able to inform practitioners without needing to run real-world screening scenarios.

\subsection{Qualitative Background}
\label{sec:Generali}

This work borrows from a previous collaborative effort at a European Fortune Global 500.
We refer to this company as G.
The purpose was to study G's hiring process as an algorithmic fairness problem.
We worked closely with Human Resources (HR), focusing on candidate screening.
In this phase of G's hiring process, an HR officer reduces the candidate pool for a job opening into a smaller pool of suitable candidates based on each candidate's profile.
The candidate pool was stored in Oracle's Taleo, a hiring platform used by HR officers to, among other things, obtain an ISO.

The following \textit{five stylized facts} summarize key practices by the HR officers (henceforth, screeners) that motivated the ISO problem.
\textit{\textbf{G1}} \textit{Varying ISOs.} Screeners chose the ISO. The choice was restricted by the sorting fields of the hiring platform, such as using the candidates' last name.
\textit{\textbf{G2}} \textit{Two ways to search the candidate pool.} Two search practices became apparent: full or partial search of the candidate pool.
\textit{\textbf{G3}} \textit{Meeting the set of minimum basic requirements.} Screeners were able to differentiate candidates relative to each other, but their focus was on finding candidates that met these requirements. 
Order within the selected $k$ candidates was not necessarily important.
\textit{\textbf{G4}} \textit{Diverse suitable candidates.} Fairness goals already existed in the form of representation quotas, often around gender, that were enforced by the screeners. 
\textit{\textbf{G5}} \textit{A consistent notion of time.} Screeners aimed at spending one minute per candidate.

Although G1-5 are specific to G, they highlight salient aspects of real-world candidate screening problems likely to hold in similar settings involving humans searching a pool of candidates (see, e.g., \cite{DBLP:journals/jcmc/PanHJLGG07, DBLP:conf/clef/GrotovCMSXR15, DBLP:conf/www/RichardsonDR07, DBLP:conf/chi/EchterhoffYM22, PeiEAAMO2023}).
G4-5 are standard to the fair set selection problem, especially under an algorithmic screener, while G1-3 introduce new considerations to such problem formulation, especially under a human screener.
We come back to G1-5 in Sections \ref{sec:ProblemFormulation} and \ref{sec:HumanScreener}.

%
%

\subsection{Related Work}
\label{sec:RelatedWork}

This paper focuses on the screener's search behavior given an ISO.
The creation of the ISO itself can be modeled as a fair set selection or, if order matters, fair ranking problem \cite{Zehlike2023_FairRanking_P1, Zehlike2023_FairRanking_P2, DBLP:journals/vldb/PitouraSK22} in which the goal is to learn a fair ISO from data using, e.g., probability-based \cite{DBLP:conf/ssdbm/YangS17, DBLP:conf/cikm/ZehlikeB0HMB17} and exposure-based \cite{DBLP:journals/cacm/Baeza-Yates18, DBLP:journals/sigir/JoachimsGPHG17} methods.
Similarly, other works, e.g., \cite{DBLP:conf/wsdm/KokkodisPI15, DBLP:conf/kdd/Kokkodis18, Hadass_2004_TheEffectInternetRecruiting} study how IAS, like online job markets, enable the creation of candidate pools and, in turn, the ISO.
Instead, motivated by our experience at G, we are interested in how a screener, particularly a human one, searches the ISO.
With some exceptions \cite{DBLP:conf/chi/EchterhoffYM22, PeiEAAMO2023, Ovaisi_10.1145/3652864, DBLP:conf/wsdm/WangGBMN18},
most fairness works avoid studying the user of the ISO, e.g., treating position bias as a technical bias.
Our screener-centric approach is similar to web search click models that were the first to formalize \cite{CraswellZTR08_ExperimentsClickPositionBias} and test \cite{DBLP:journals/jcmc/PanHJLGG07, DBLP:conf/clef/GrotovCMSXR15, DBLP:conf/www/RichardsonDR07} how users search over an ISO.
Different from these works, we consider a user that ``clicks'' on more than one item, and formalize such user under a utility-maximizing framework with fairness constraints, relating the insights from these works to candidate screening as well as to fair set selection works \cite{stoyanovich2018online}.

The works on click models provide empirical evidence for the position bias, though precede the fairness literature.
Two recent works provide empirical evidence for position bias in candidate screening due the ISO with a focus on individual fairness \cite{DBLP:conf/innovations/DworkHPRZ12}.
\citet{DBLP:conf/chi/EchterhoffYM22} collaborate with a college to study anchoring bias \cite{tversky_judgment_1974} in admissions officers. 
They find that the same applicant is better off if it is preceded by worst rather than better applicants, and propose an algorithm to balance out the anchoring bias when presenting applications to the admissions officer.
\citet{PeiEAAMO2023} also collaborate with a college to study how the platform used by professors for evaluating homeworks affects student grades. 
They run experiments varying the order in which the homeworks are presented by the platform, and show that the default alphabetical order unfairly rewards students with the same work quality due to their last names.
We differ from these works by defining the ISO as a parameter in the problem formulation and using simulations flexible enough to capture these and other screening scenarios.

Our best-$k$ and good-$k$ formulations (Section~\ref{sec:ProblemFormulation.Objectives}) belong to the fair set selection literature \cite{DBLP:conf/eaamo/BueningSBGD22}. 
The reference problem is the secretary problem (SP) \cite{ferguson1989solved} where we select a candidate in a randomly ordered sequence committing to the acceptance or rejection decision after each evaluation.
The past literature has already analyzed the fairness implication of the SP, even in the $k$-choice extension \cite{stoyanovich2018online}.
However, our best-$k$ and good-$k$ formulations, differently from the SP, focus on the screening process, not the interview phase. 
We assume an offline set selection, meaning each candidate is individually evaluated by the screener and each decision is irrevocable. 
Such setting emphasizes the ISO's role.
Past works on the SP online setting assume the additive utility we employ in the best-$k$ formulation \cite{stoyanovich2018online, mehrotra2021mitigating}.
This fact shows how peculiar our good-$k$ formulation is relative to previously analyzed settings. 

%
%

%
%

\section{Searching the Pool of Candidates}
\label{sec:ProblemFormulation}

We formulate the set selection problem, in which a decision-maker selects a set of items from a population, given an ISO.
Here, the candidates for a job represent the items and the screener evaluating their profiles represents the decision-maker.
Let the ISO be the product of an IAS specific to hiring.

\subsection{Setting}
\label{sec:ProblemFormulation.Setting}

Let us consider a \textit{candidate pool} $\mathcal{C}$ of $n$ candidates, where each \textit{candidate} $c$ is described by the \textit{vector of $p$ attributes} $\mathbf{X}_c \in \mathbb{R}^{p}$ and the \textit{protected attribute} $W_c$.
We assume that $W$ is binary, such that $W_c = 1$ if $c$ belongs to the protected group and $W_c = 0$ otherwise; we can relax this assumption if needed.
The candidates are evaluated by a \textit{screener} $h \in \mathcal{H}$, where $\mathcal{H}$ denotes the set of available screeners. 
The following variables refer to a specific $h$. 
The goal of $h$ is to obtain a \textit{set of $k$ selected candidates} $S^k \in [\mathcal{C}]^k$, with $[\mathcal{C}]^k$ denoting the set of $k$-subsets of $\mathcal{C}$, 
based on each candidate's application profile as summarized by the tuple $(\mathbf{X}_c, W_c)$.
Candidate evaluation occurs when $h$ uses an \textit{individual scoring function} $s \colon \mathbb{R}^{p} \to [0, 1]$, such that $s(\mathbf{X}_c)$ returns the score of $c$, and $h$ cannot use $W_c$ when scoring $c$. 
The higher the score, the better $c$ fits the job.

The screener $h$ explores the candidate pool $\mathcal{C}$ in a specific order. 
We denote the \textit{set of total orderings of candidates} in $\candidatesset$ by $\Theta$.
An \textit{order} $\sigma \in \Theta$ maps an integer $i \in \{1, \dots n\}$ to a candidate $c \in \candidatesset$, indicating that $c$ occupies the $i$-th position according to $\sigma$, with notation $\sigma(i) = c$ and vice-versa $\sigma^{-1}(c) = i$.
Importantly, the screener explores $\mathcal{C}$ under the ISO $\theta \in \Theta$, which represents the order chosen by or, alternatively, provided to $h$ before searching $\candidatesset$ (recall, G1 in Section~\ref{sec:Generali}) via an IAS. 
The screener is not required to explore the entirety of $\mathcal{C}$, meaning $h$ can either fully or partially explore $\candidatesset$ given $\theta$ (recall, G2 in Section~\ref{sec:Generali}).
We assume that the screener \textit{respects} $\theta$, meaning: 
\begin{equation}
\label{eq:order}
    \mbox{$c_1 \in \candidatesset$ is evaluated before $c_2 \in \candidatesset$ only if $\theta^{-1}(c_1) < \theta^{-1}(c_2)$.}
\end{equation}

\subsection{Two Problem Formulations}
\label{sec:ProblemFormulation.Objectives}

We formulate two utility-based set selection problems for $h$ with the shared objective of achieving an optimal and fair set $S^k$.
Under the \textit{best-}$k$ formulation, $S^k$ represents \textit{the fair best $k$ candidates} in $\candidatesset$ according to $h$; we denote it as $S^k_{\besttext}$. 
Under the \textit{good-}$k$ formulation, $S^k$ represents \textit{the fair first good-enough $k$ candidates} in $\candidatesset$ according to $h$; we denote it as $S^k_{\goodtext}$.
The key difference between the two is that the best-$k$ requires a full search of $\candidatesset$ while the good-$k$ allows for a partial search of $\candidatesset$ under $\theta$.

How we define optimality, as shown in Sections~\ref{sec:best-k} and \ref{sec:good-k}, determines the best-$k$ and good-$k$ problems.
For fairness, we define the \textit{representational quota} $q \in [0, 1]$ as the desired fraction of protected candidates in $S^k$ and use the fraction $f\big( S^k \big) \in [0, 1]$:
\begin{equation}
\label{eq:fairness_function}
    f\big( S^k \big) = \frac{\left\vert\{c \in S^k \text{ s.t. } W_c = 1\}\right\vert}{k}
\end{equation}
for $h$ to meet $q$ when deriving $S^k$ by satisfying the condition $f\big( S^k \big) \geq q$.
The unconstrained version is achieved by $q=0$.
We view $q$ as a policy enforced by $h$ to achieve a diverse $S^k$ (recall, G4 in Section~\ref{sec:Generali}). 
It is a statement on the composition of $S^k$, not a statement on the ordering of protected candidates within $S^k$.\footnote{For $k=10$ and $q=0.5$, e.g., the fair screener would need to derive $S^k$ with $5$ protected candidates though in no particular order within $S^k$.}

\subsubsection{Best-$k$.}
\label{sec:best-k}

The screener $h$ finds the set of best $k$ candidates in $\candidatesset$ given $q$ while respecting the ISO \eqref{eq:order}.
Here, $h$ needs to evaluate the complete $\candidatesset$ since it must score and rank all candidates according to the individual scoring function $s$ before choosing the ones with the highest scores and that satisfy $q$.

We view the goal in terms of maximizing a utility for $h$. 
We define \textit{utility} as the benefit derived by $h$ from selecting $k$ candidates. 
Formally, utility is a function $U^k \colon [\mathcal{C}]^k \, \times \, \Theta \to \mathbb{R}$. 
The simplest expression for $U^k$ is to add the scores of the selected candidates:
\begin{equation}
\label{eq:Utility}
    U^k_{\addtext} \big( S^k, \theta \big) = \sum_{c \in S^{k}} s\big( \mathbf{X}_{c} \big)
\end{equation}
rationalizing that $h$ maximizes its utility by selecting the $k$ most suitable candidates given $\theta$. 
Notice that $\theta$ in \eqref{eq:Utility} does not affect the evaluation order of $S^k$ due to the commutative property of addition.
Under \eqref{eq:Utility}, we define  \textbf{the best-\textit{k} problem} as:
\begin{equation}
\label{eq:fair_objective_all_screener}
    \begin{aligned}
    \argmax_{S^k \in [\mathcal{C}]^k} & \quad U^k_{\addtext} \big(S^k, \theta\big) \\
    \textrm{s. t.} & \quad f(S^k) \geq q
    \end{aligned}
\end{equation}
with its solution as $S^k_{\besttext}$. In the presence of tied scores, $S^k_{\besttext}$ may not be unique. In such a case, we consider any solution.
We emphasize that \eqref{eq:Utility} is not the only possible model for the utility of $h$ and alternative models, such as exposure discounting \cite{DBLP:conf/kdd/SinghJ18}, can be considered for \eqref{eq:fair_objective_all_screener}.
We leave this for future work.

\begin{figure*}[t]
\scalebox{0.85}{
\begin{minipage}[t]{0.425\textwidth}
\begin{algorithm}[H]
\caption{ExaminationSearch}
\label{algo:Examination}
\begin{algorithmic}[1]
\Require $n$, $\theta$, $k$, $q$
\Ensure $S^k_\besttext$
\State $q^* \, \gets$ \texttt{round(}$q \cdot k$\texttt{)}; $r^* \, \gets k - q^*$
\State $\mathit{scores} \, \gets [ s(\mathbf{X}_{c})\ \mathit{for}\ c = \theta(1), \ldots, \theta(n) ]$
\State $\tau \gets \texttt{argsortdesc(} \mathit{scores}$\texttt{)}
\State $i \, \gets 1$; $k^* \, \gets 0$; $Q \, \gets \{ \}$; $R \, \gets \{ \}$
\While{$k^* < k$} 
    \State $c \; \gets \theta(\tau(i))$; $i \, \gets i + 1$
    \If{$W_c == 0$ \textbf{and} \texttt{len(}$R$\texttt{)}$== r^*$ }
       \State \textbf{continue}
    \EndIf
    \State $k^* \, \gets k^* + 1$
        \If{$W_c==1$ \textbf{and} $\texttt{len(}Q\texttt{)} < q^*$}
             \State $Q \, \gets Q \cup \{ c \}$
        \Else
             \State $R \, \gets R \cup \{ c \}$
       \EndIf
\EndWhile
\Return $Q \cup R$
\end{algorithmic}
\end{algorithm}
\end{minipage}
}
\hspace{0.08\textwidth}
\scalebox{0.85}{
\begin{minipage}[t]{0.425\textwidth}
\begin{algorithm}[H]
\caption{CascadeSearch}
\label{algo:Cascade}
\begin{algorithmic}[1]
\Require $n$, $\theta$, $k$, $q$, $\psi$
\Ensure $S_{\goodtext}^k$
\State $q^* \, \gets$ \texttt{round(}$q \cdot k$\texttt{)}; $r^* \, \gets k - q^*$
\State $i \, \gets 1$; $k^* \, \gets 0$; $Q \, \gets \{ \}$; $R \, \gets \{ \}$
\While{$k^* < k$} 
    \State $c \; \gets \theta(i)$; $i \, \gets i + 1$
    \If{$W_c == 0$ \textbf{and} \texttt{len(}$R$\texttt{)}$== r^*$ }
       \State \textbf{continue}
    \EndIf
    \State $Y_c \; \gets s(\mathbf{X}_c)$ 
    \If{$Y_c \geq \psi$}
      \State $k^* \, \gets k^* + 1$
        \If{$W_c==1$ \textbf{and} $\texttt{len(}Q\texttt{)} < q^*$}
             \State $Q \, \gets Q \cup \{ c \}$
        \Else
             \State $R \, \gets R \cup \{ c \}$
       \EndIf
      \EndIf
\EndWhile
\Return $Q \cup R$
\end{algorithmic}
\end{algorithm}
\end{minipage}
}
\caption{Search procedures, respectively, for the best-$k$ and good-$k$ problems (Section~\ref{sec:ProblemFormulation}). Each represents the algorithmic screener, $h_a$, of each problem due to a lack of fatigue (Section~\ref{sec:HumanScreener}). Further, for the human-like screener, $h_h$, we obtain the \textit{HumanLikeExaminationSearch} and the \textit{HumanLikeCascadeSearch} (Algorithms 3 and 4 in the Appendix)
by further requiring $\Phi$ and adding, respectively, $\epsilon(\Phi(t-1))$ to line 2 and line 7 when computing the individual scores while tracking time $t$.}
\label{fig:TheAlgos}
\end{figure*}

\subsubsection{Good-$k$}
\label{sec:good-k}

The screener $h$ finds $k$ candidates in $\candidatesset$ that meet a set of \textit{minimum basic requirements} $\psi$ (recall, G3 in Section~\ref{sec:Generali}) given $q$ while respecting the ISO \eqref{eq:order}.
We represent $\psi$ as a \textit{minimum score}, such that $h$ deems a candidate $c \in \candidatesset$ as eligible for being selected if $s(\mathbf{X}_c) \geq \psi$.
Unlike the best-$k$ formulation, here $h$ is not required to evaluate the whole $\candidatesset$ as it is enough to find the first $k$ candidates that are good enough according to $\psi$ and that satisfy $q$. 

We still view the goal in terms of maximizing a utility for $h$. We need to, however, define an alternative utility function to \eqref{eq:Utility} that ensures $h$ stops searching $\candidatesset$ after finding the $k$-\textit{th} good-enough candidate according to $\psi$.
We define the following expression:
\begin{equation}
\label{eq:AlternativeUtility}
    U^k_{\psi}\big( S^k, \theta \big) = \left\{
    \begin{array}{ll}
        k - \sum_{c \in S^k} p(c, S^k, \theta) & \text{if,} \  \forall c \in S^k, \  s(\mathbf{X}_c) \geq \psi   \\
        0 & \text{otherwise.}
    \end{array} \right.
\end{equation}
with the \textit{penalty function} defined as:
\begin{equation}
\label{eq:Penalty}
\begin{aligned}
    p(c, S^k, \theta) = & \, \mathbb{1} \big \{ \exists\ c' \in \mathcal{C}\setminus S^k \, \\ & \mbox{s.t.}\ \, \theta^{-1}(c') < \theta^{-1}(c) \wedge s(\mathbf{X}_{c'}) \geq \psi \wedge W_{c'}=W_c \big \}.
\end{aligned}
\end{equation}
Under \eqref{eq:AlternativeUtility}, $h$ wants to find as quickly as possible the $k\text{-\textit{th}}$ suitable candidate without wanting to check whether the $(k+1)\text{-\textit{th}}$ candidate is also suitable.
This is because, for a candidate $c$, \eqref{eq:Penalty} looks for another candidate of the same group as $c$ and meeting $\psi$, who occurs before $c$ under $\theta$ but who has not been selected into $S^k$.
It models the ``wasted effort" in choosing a candidate occurring after another one meeting all the same requirements. 
At worst, there are $k$ penalties.
Under \eqref{eq:AlternativeUtility}, we define \textbf{the good-\textit{k} problem} as:
\begin{equation}
\label{eq:fair_objective_U_psi}
    \begin{aligned}
    \argmax_{S^k \in [\mathcal{C}]^k} & \quad U^k_{\psi} \big(S^k, \theta\big) \\
    \textrm{s. t.} & \quad f(S^k) \geq q
    \end{aligned}
\end{equation}
with its solution as $S_{\goodtext}^k(\psi)$ or, if there is no ambiguity on $\psi$, simply as $S_{\goodtext}^k$.
When the fairness constraint is strengthened to a fixed quota, $f(S^k) = q$, the solution is unique; in the general case, $f(S^k) \geq q$, there can be two solutions but with different fractions of the protected group.
Similarly to the best-\textit{k} problem, we emphasize that \eqref{eq:AlternativeUtility} is not the only utility model for \eqref{eq:fair_objective_U_psi}; other models are possible as long as they describe the partial search. 

%
\begin{remark}
\label{remark:ISOandPS}
    $\theta$ influences the screening process under the good-$k$ problem \eqref{eq:fair_objective_U_psi} due to the potential partial search of $\mathcal{C}$ by $h$, affecting which $k$ candidates are selected.
\end{remark}
To observe Remark~\ref{remark:ISOandPS}, 
let $k=1$ and assume two candidates such that $s(\mathbf{X}_{c_1}) \geq \psi$ and $s(\mathbf{X}_{c_2}) \geq \psi$. A $\theta$ such that $\theta^{-1}(c_1) = 1$ and $\theta^{-1}(c_2) = 2$ would imply that $c_1$ is considered eligible and is selected before $h$ even evaluates $c_2$. 
Conversely, a reverse $\theta$ such that $\theta^{-1}(c_1) = 2$ and $\theta^{-1}(c_2) = 1$ would imply the opposite.
This remark holds without assuming anything about $h$.

\subsection{Two Search Procedures}
\label{sec:ProblemFormulation.Algorithms}

The \textit{ExaminationSearch} 
(Algorithm~\ref{algo:Examination}) solves the best-$k$ problem, returning $S^k_\besttext$ for given $n$ (candidates) and $\theta$ (ISO), and parameters $k$ (subset size) and $q$ (group fairness constraint).
First, line 2 calculates the minimum number $q^*$ of candidates from the protected group to be selected, and the maximum number of candidates $r^*$ not in that quota. 
Then, candidates are considered by descending scores, using the \texttt{argsortdesc} procedure (lines 2-3). 
The loop in lines 5-13 iterates until $k$ candidates are found. 
The loop adds candidates to the sets $Q$ and $R$: $Q$ are candidates in the quota of the protected group; $R$ are candidates not in that quota that can be non-protected or protected. 
A non-protected candidate can be only added to the $R$ set; thus, line 7 checks if there is still room in $R$ to do this. A protected candidate is added to the quota set $Q$ if there is room (lines 10-11) or to the other set $R$ otherwise (lines 12-13). 
Finally, the procedure returns the candidates in the quota set $Q$ or in the other set $R$. 
The result of the \textit{ExaminationSearch} procedure maximizes (\ref{eq:fair_objective_all_screener}), as candidates are added in decreasing score, while keeping the fairness constraint through the quota management.

The \textit{CascadeSearch} 
(Algorithm~\ref{algo:Cascade}) solves the good-$k$ setting, returning $S^k_\goodtext$ for given $n$ and $\theta$, and parameters $k$, $q$ and $\psi$ (min. basic requirement).
The difference with the \textit{ExaminationSearch} procedure consists in strictly following $\theta$ (line~4) and checking $\psi$ (line~8) before adding a candidate to the quota set $Q$ or to the other set $R$.
The result of the \textit{CascadeSearch} procedure maximizes (\ref{eq:fair_objective_U_psi}), as no penalty is accumulated in the loop. 
This is because, a non-protected candidate ($W_c=0$) is not added only if there is no room in $R$ (and $R$ never gets smaller to allow for more room later on), while a protected candidate ($W_c=1$) is not added only if it does not meet $\psi$ in line 8 and, thus it cannot be counted for the penalty. 

Our aim here is not to provide novel optimal algorithms, but to study the screener's search behavior of $\candidatesset$ under $\theta$.
Hence, we move away from an optimality analysis of the two algorithms (e.g., \cite{fagin2001optimal}), and focus on modeling the search behavior of $h$ when solving the two problems.
Both algorithms are inherently sequential and can be applied online because we aim to model a human-like screener (next section) that operates sequentially.
Yet, in both algorithms, the applicants are disclosed according to $\theta$ and not to the score, differently from past set selection works (e.g., \cite{stoyanovich2018online}).

%
%

\section{The Human-Like Screener}
\label{sec:HumanScreener}

To study the human interaction with an ISO, we distinguish two kinds of screeners based on the proneness to error when evaluating candidates: 
$h$ is an \textit{algorithmic screener}, or $h_a \in \mathcal{H}_a$, if it consistently evaluates $\mathcal{C}$; whereas $h$ is a \textit{human-like screener}, or $h_h \in \mathcal{H}_h$, if its fatigue hinders the consistency of its evaluation of $\mathcal{C}$.

\subsection{Fatigue and Fatigued Scores}
\label{sec:HumanScreener.BiasedScores}

We first introduce a \textit{time component}, with $t$ denoting the discrete unit of time that represents how long $h$ takes to evaluate a candidate $c \in \mathcal{C}$. 
We assume that $t$ is constant (recall G5 Section~\ref{sec:Generali}), implying that time cannot be optimized by $h$.
We track time along $\theta$, meaning $h$ evaluates the first candidate that appears in $\theta$ at time $t=1$ and so on. 
Time $t$ ranges from $0$ to $n$ at maximum.
We then introduce a \textit{fatigue component} $\phi(t)$ specific to $h_h$ as a function of $t$ and model the \textit{accumulated fatigue} $\Phi \colon \{0, \ldots, n\} \to \mathbb{R}$, with $\Phi(0) = 0$. 
The discrete derivative of $\Phi$, i.e., $\phi(t) = \Phi(t) - \Phi(t-1)$, defined for $t \geq 1$, is the effort of $h_h$ to examine the $t$-th candidate. 

How we define $\Phi$ conditions the effect of fatigue on our analysis of $h_h$.
We make the simplest modeling choice for $\phi$ by assuming that \textit{fatigue accumulates linearly over time}, or $\phi(t) = \lambda$ so that $\Phi(t) = \lambda \cdot t$, meaning $h_h$ becomes tired over time at a constant pace.
How does fatigue materialize for $h_h$?
We model the effect of fatigue on $h_h$ through the \textit{fatigued score}:
\begin{equation}
\label{eq:BiasedScoresForHh}
    s_{h_h}(\mathbf{X}_c) + \epsilon
\end{equation}
where $\epsilon$ is a random variable dependent on $\Phi$ that quantifies the deviation from the \textit{truthful score} $s_h(\mathbf{X}_c)$.

We model $\epsilon$ using \emph{two modeling choices} at a given time $t$.
    \textit{\textbf{First modeling choice}:}
    $\epsilon_1$ is a centered Gaussian, and the fatigue affects only its variance. 
    Formally, $\epsilon_1 \sim \mathcal{N}(0, \, v(\Phi(t-1)))$, where $v \colon \mathbb{R} \to \mathbb{R}$ defines the variance of $\epsilon_1$ as an increasing function of $\Phi$.
    \textit{\textbf{Second modeling choice}:}
    $\epsilon_2$ as an uncentered Gaussian, whose mean is a decreasing function of the fatigue.
    Formally, $\epsilon_2 \sim \mathcal{N}(\mu(\Phi(t-1)), \, v(\Phi(t-1))$, where $\mu \colon \mathbb{R} \to \mathbb{R}$ is a decreasing function rather than a constant of $\Phi$.
 
Under $\epsilon_1$, $h_h$ tends to overscore or underscore candidates, introducing both negative and positive bias (i.e., fatigue as ``less attention'' when evaluating candidates) over time. 
Under $\epsilon_2$, instead, $h_h$ tends to underscore the candidates (i.e., fatigue as ``less effort'' when evaluating candidates) over time, introducing always a negative bias. 
With $\epsilon_1$ and $\epsilon_2$, we capture two realistic biased settings driven by the ISO.
We assume $h_h$ is unaware of its fatigue, representing an unconscious bias from performing a repetitive task \cite{Kahneman2011Thinking, Kahneman2021Noise}.

\begin{remark}
\label{remark:HumanAndIF}
    $\Phi$ implies that $h_h$ evaluates identical candidates $c_1$ and $c_2$ differently under $\theta$ at $t_1$ and $t_2$ as long as $\Phi(t_1) \neq \Phi(t_2)$ and regardless of whether $h_h$ solves for the best-$k$ or good-$k$ problem.
\end{remark}

The algorithms~\ref{algo:Examination} and \ref{algo:Cascade} represent $h_a$.
To represent the human-like screener $h_h$, we must track $\Phi$ over time and draw $\epsilon_1$ (or $\epsilon_2$) to compute \eqref{eq:BiasedScoresForHh} of $h_h$ at time $t$. 
The only changes are to line 2 in Algorithm~\ref{algo:Examination} and line 7 in Algorithm~\ref{algo:Cascade}, where the score computed for candidate $c$ is biased by $\epsilon_1$ (or $\epsilon_2$). 
We present the human-like versions of these search procedures as Algorithms 3 and 4 in Appendix;
these algorithms, though, can be observed using Figure~\ref{fig:TheAlgos}.

\subsection{Position Bias Implications}
\label{sec:PositionBias}

We analyze the fairness and optimality implications of the position bias implicit to $\theta$. For concreteness, we make \textit{two assumptions}.
\textit{\textbf{A1}: We assume $\theta$ is independent of the protected attribute}, meaning how candidates appear in $\theta$ contains no information about $W$.
\textit{\textbf{A2}: We assume the individual scoring function evaluates candidates fairly and truthfully}, meaning $s(\mathbf{X}_c)$ captures no information about $W_c$.
Under \textit{A1} and \textit{A2}, we control for other biases, such as measurement error in $s$, and focus on the position bias coming from $\theta$.

We start with the fairness implications for the best-$k$ and good-$k$ problems. 
Given $\theta$, it is important to distinguish between the group fairness constraining $h$ (i.e., the quota $q$) and the individual fairness violation when $h$ fails to evaluate similar candidates similarly.
Regarding group-level fairness, both $h_a$ and $h_h$ are fair in solving for \eqref{eq:fair_objective_all_screener} and \eqref{eq:AlternativeUtility} by satisfying $f\big(S^k\big) \geq q$.
This point is clear for $h_a$ in Algorithms \ref{algo:Examination} and \ref{algo:Cascade} as there is no fatigue involved.
The same holds for $h_h$ in Algorithms~3 and 4 because the error on the score does not affect the evaluation of $q$.
Here, we have that the expected error $\mathbb{E}[\epsilon \mid W_c = 1, \theta] = \mathbb{E}[\epsilon \mid W_c = 0, \theta]$, regardless of $\epsilon_1$ or $\epsilon_2$ for $h_h$. 
The fatigue and, thus, the fatigued scores are, on average, shared across protected and non-protected candidates.

Distinguishing between $h_a$ and $h_h$ becomes important under individual-level fairness because $h_a$ ensures it while $h_h$ violates it.
A candidate's position in $\theta$ influences the amount of error made by $h_h$ when evaluating that candidate. 
Similar candidates will not be evaluated similarly due to the unequal accumulation of fatigue experienced by $h_h$ when searching $\theta$.
For $\epsilon_2$, e.g., even if $\mathbf{X}_{c} = \mathbf{X}_{c'}$ but $\theta^{-1}(c) > \theta^{-1}(c')$, the score of $c$ is less, on average, than the one of $c'$, and $c'$ has an unfair premium over $c$ from $\theta$.

We now consider the optimality implications for the best-$k$ and good-$k$ problems.
It follows that $h_a$ reaches the optimal solution in both problems as the absence of fatigue enables $h_a$ to consistently judge suitable candidates.
The opposite holds for $h_h$ due to the inconsistent scoring of candidates ascribed by the accumulated fatigue.
The biased scores not only violate individual fairness, but also lead $h_h$ to misjudge candidates, eventually choosing the wrong ones when searching $\theta$.
%
Significantly, under the $h_h$, the position of a candidate in $\theta$ matters.
As no two candidates occupy the same position in $\theta$, it impacts whether a candidate, depending on the search strategy, is evaluated or not (Remark~\ref{remark:ISOandPS}) and, if so, is evaluated fairly or not (Remark~\ref{remark:HumanAndIF}). These results only worsen when relaxing \textit{A1} and \textit{A2}. We explore these insights in Section~\ref{sec:Experiments} by considering multiple hiring settings for $h_a$ and $h_h$.

%
%

\section{An Experimental Framework}
\label{sec:Experiments}

We introduce a flexible experimental framework in R \cite{Rlang} based on Monte Carlo simulations. 
It can handle different screening scenarios under the algorithmic or human-like screeners given an ISO.
The code is available in a dedicated GitHub repository.\footnote{\url{https://github.com/cc-jalvarez/initial-screening-order-problem}}

\begin{figure*}[t]
  \centering
  \includegraphics[width=0.32\textwidth]{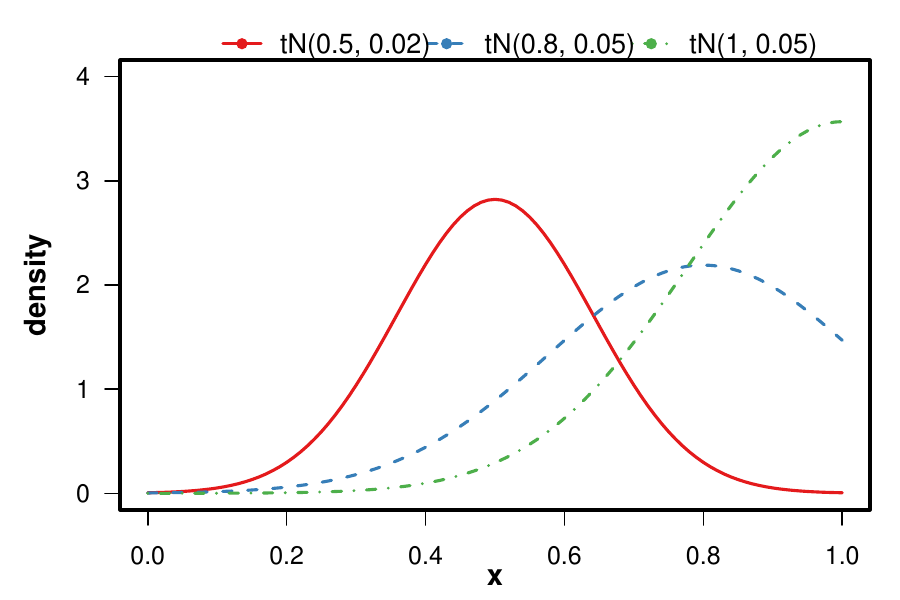}
  \hfill
  \includegraphics[width=0.32\textwidth]{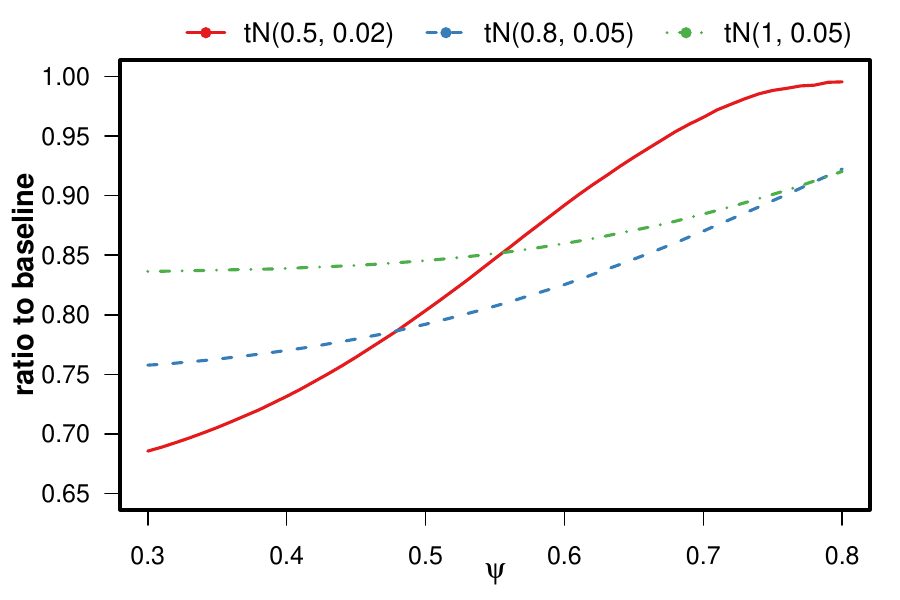}
  \hfill
  \includegraphics[width=0.32\textwidth]{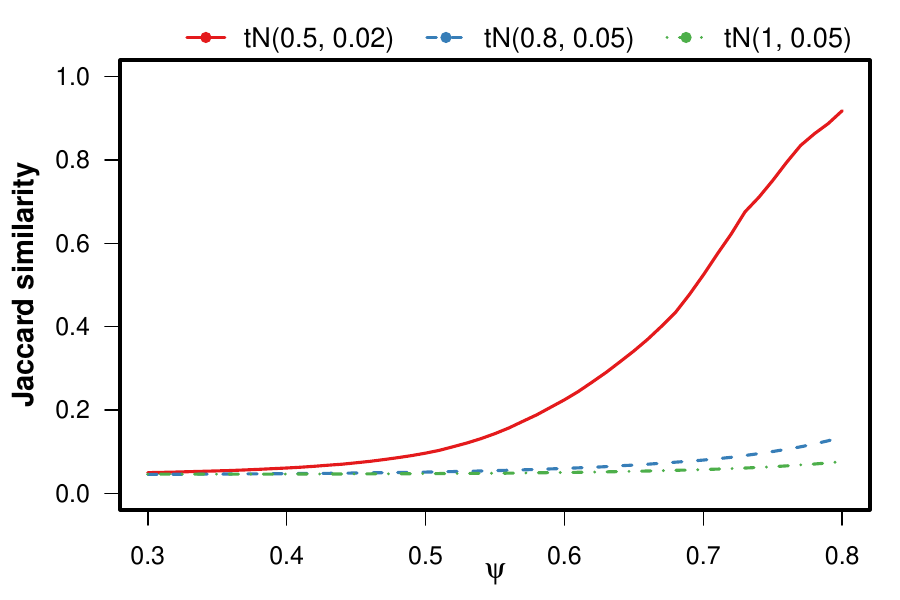} 
  \includegraphics[width=0.32\textwidth]{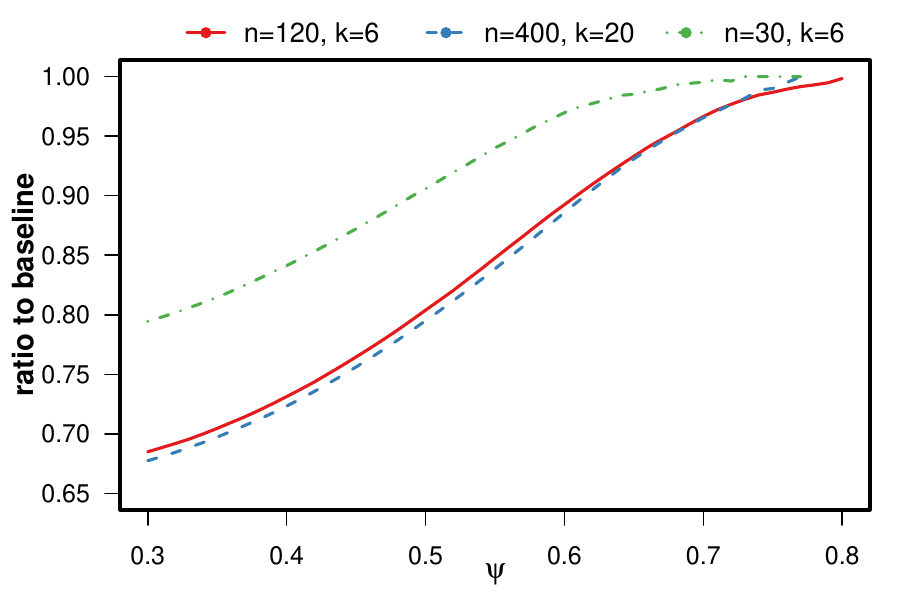}
  \hfill
  \includegraphics[width=0.32\textwidth]{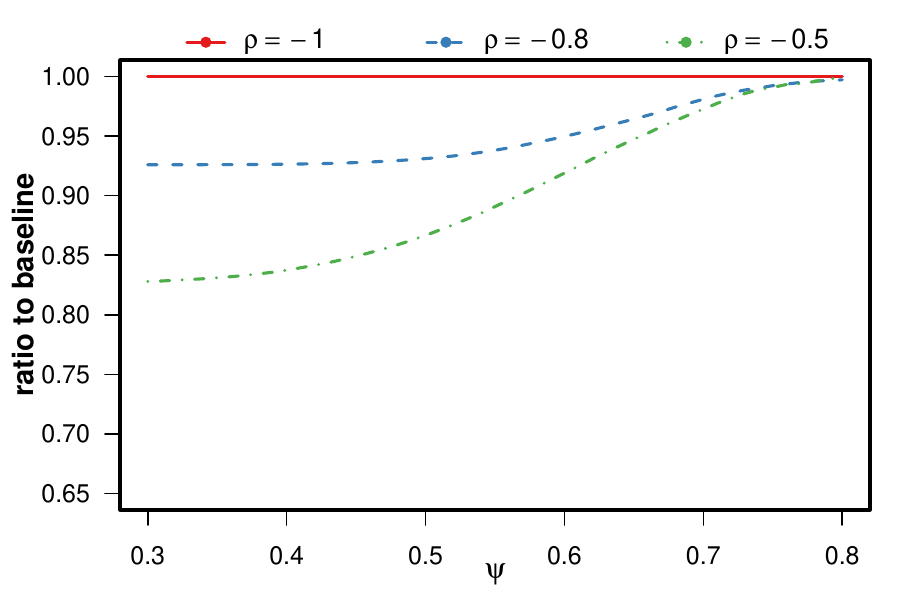}
  \hfill
  \includegraphics[width=0.32\textwidth]{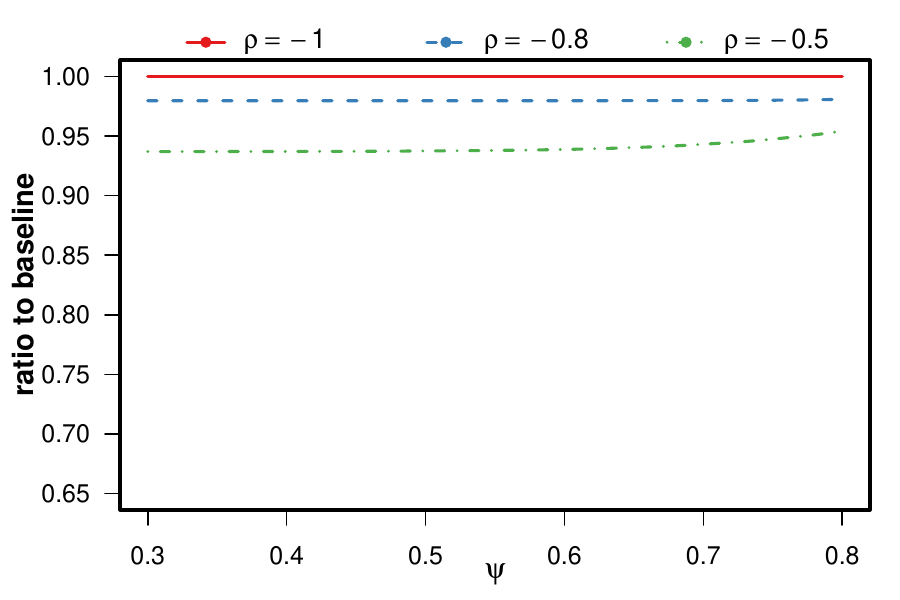} 
  \caption{
  \textit{Upper}. 
  Left: densities of candidate score distributions, with scores raging from 0 to 1.
  Center and Right: resp., RtB and JdS for all score distributions over $\psi$ under $\theta \ci s$.
  \textit{Lower.}
  Left: $n, k$ combinations for $tN(0.5, 0.02)$ distribution over $\Psi$ under $\theta \ci s$.
  Center and Right: resp., RtB under different $\rho$'s (i.e., under $\theta \not\!\perp\!\!\!\perp s$) for $tN(0.5, 0.02)$ and $tN(1, 0.05)$ distributions over $\Psi$.
  }
\label{fig:1}
\end{figure*}

\subsection{Setup}
\label{sec:Experiments.Setup}

\subsubsection{Generating the sample.}
We assume a sample consisting of $n$ triplets $\{ (s(\mathbf{X}_{c_i}), \theta({c_i}), W_{c_i}) \}_{i=1}^n$ drawn from a probability distribution with domain $\mathcal{G}_n \times \mathbb{R}^n \times \{0, 1\}^n$, where $\mathcal{G}_n$ is the set of all permutations of $\{1, \ldots, n\}$. Each sample represents a candidate pool sorted according to an ISO.

For the individual candidate scores, we consider \textit{three distributions} to model scenarios in which top candidates, as in top scores, occur with different probabilities in $\candidatesset$. 
All three distributions use the truncated normal, $tN(\mu, \sigma)$ \cite{Botev2017}, with values bounded in $[0, 1]$. 
These scenarios, shown in Figure~\ref{fig:1} (upper, left), are:
\begin{itemize}
    \item \textit{Symmetric distribution of scores} (in red) defined by $\mu = 0.5$ and $\sigma = 0.02$, implying that top candidates occur with a very low probability in $\candidatesset$.
    \item \textit{Asymmetric distribution of scores} (in blue) defined by $\mu = 0.8$ and $\sigma = 0.05$, implying that top candidates occur with a higher probability and median value ($\approx 0.75$) in $\candidatesset$ relative to the previous scenario.
    \item \textit{Increasing distribution of scores} (in green) defined by $\mu = 1$ and $\sigma = 0.05$, implying that top candidates occur with an even higher probability and median value ($\approx 0.85$) in $\candidatesset$ relative to the two previous scenarios.
\end{itemize}
These scenarios have implications, in particular, for the good-$k$ problem where we set the minimum score $\psi$ and the screener is not required to explore all of $\candidatesset$ under $\theta$.
A large $\psi$ makes the screening process highly selective in the first scenario, less selective in the second scenario, and not selective at all in the third scenario, representing candidate pools with different candidate quality.

For the ISO, we consider \textit{two settings} in which a $\theta$ relates or not to $s(\mathbf{X}_{c_i})$, allowing to explore $\theta$ as a product of different IAS:
\begin{itemize}
    \item \textit{$\theta$ is generated randomly and independently from the individual candidate scores}. Formally, $\theta \ci s$.
    \item \textit{$\theta$ is generated randomly with a correlation $\rho$ with the individual candidate scores}, where $\rho$ is the Spearman's rank correlation of the pairs $\{ (\theta(c_i), s(\mathbf{X}_{c_i})) \}_{i=1}^k$.\footnote{To generate the correlated $\theta$, we rely on copulas \cite[Section 3.4]{EMBRECHTS2003329}.} Formally, $\theta \not\!\perp\!\!\!\perp s$.
\end{itemize}
Intuitively, under $\theta \ci s$, $\theta$ carries no information about the candidate quality in $\candidatesset$. It captures settings in which the screener sorts alphabetically or performs a random shuffle of the candidates.
Under $\theta \not\!\perp\!\!\!\perp s$, in particular for $\rho=-1$, $\theta$ sorts candidates by descending scores. It captures settings where the screener obtains a ranked list of candidates from the IAS.

Finally, regarding the protected attribute, we consider a sample of candidates drawn from $Ber(\mathit{pr})$ such that $\mathit{pr}=0.2$ is the fraction of protected candidates, i.e., $W_c=1$, in $\candidatesset$.
The sample is independently drawn both from the scores and ISO, following assumptions \textit{A1} and \textit{A2} in Section~\ref{sec:PositionBias}.
We can increase $\mathit{pr}$ to study a more diverse $\candidatesset$ and its effect on the screener reaching $q$.

For each set of the parameters ($n$, $k$, $q$, $\rho$, $\psi$), we run 10,000 times the experiments by randomly generating $n$ triplets at each run.
We consider $n=120, k=6, q=0.5$ for Sections~\ref{sec:Experiments.Metrics.outFatigue} and \ref{sec:Experiments.Metrics.withFatigue}.
The runs without a solution of the problem are discarded. 
This mainly occurs in the good-$k$ problem when there are not enough $k$ candidates with scores greater or equal than $\psi$.

\begin{figure*}[t]
  \centering
  \includegraphics[width=0.32\textwidth]{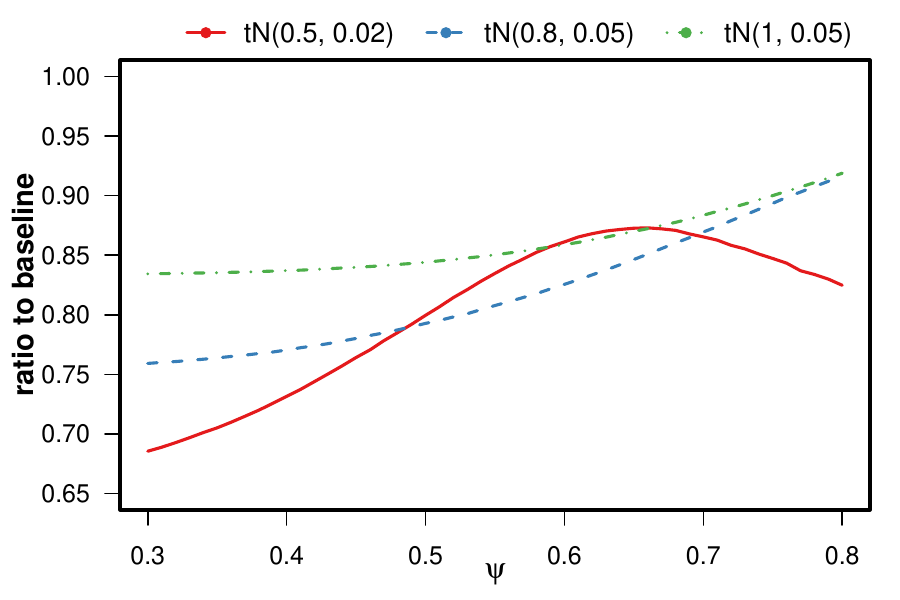}
  \hfill
  \includegraphics[width=0.32\textwidth]{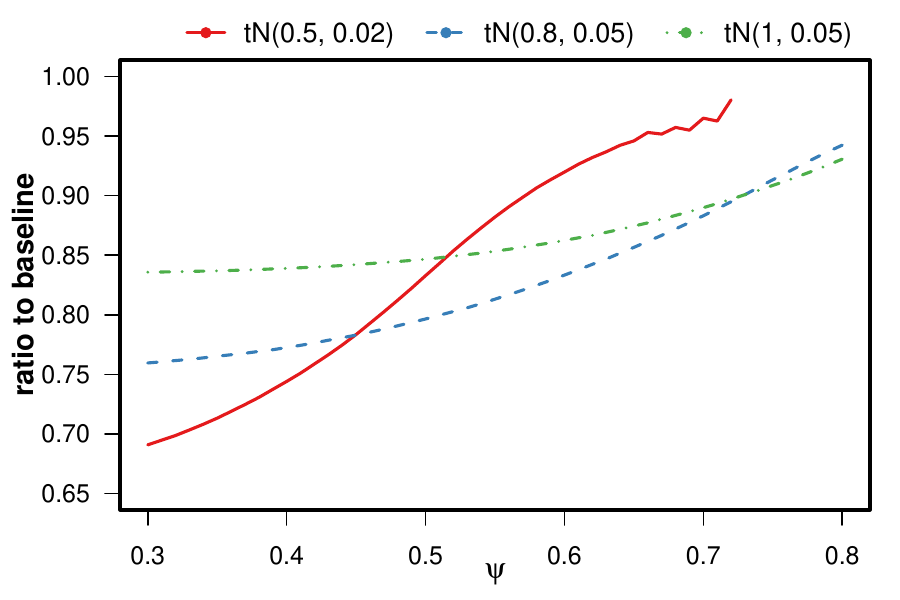}
  \hfill
  \includegraphics[width=0.32\textwidth]{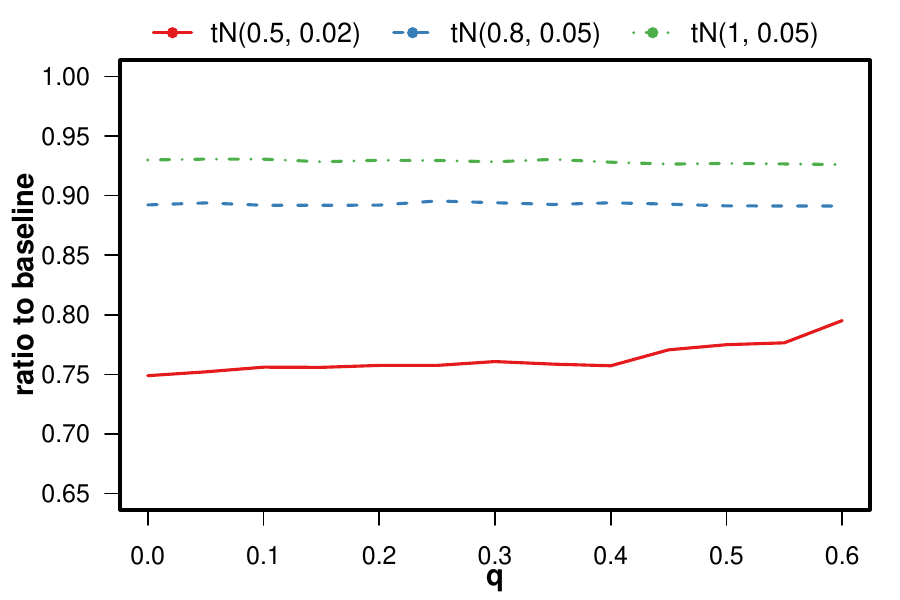} 
  \includegraphics[width=0.32\textwidth]{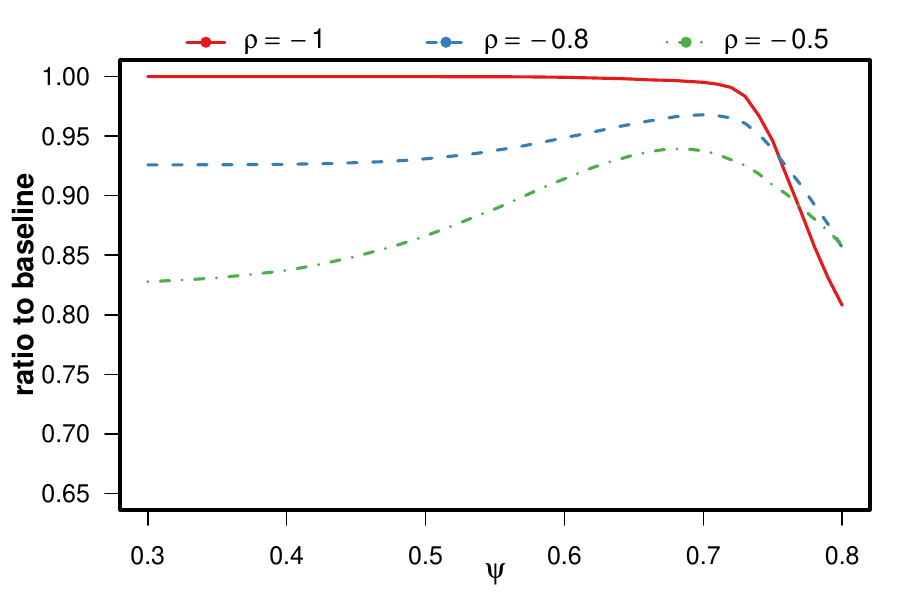}
  \hfill 
  \includegraphics[width=0.32\textwidth]{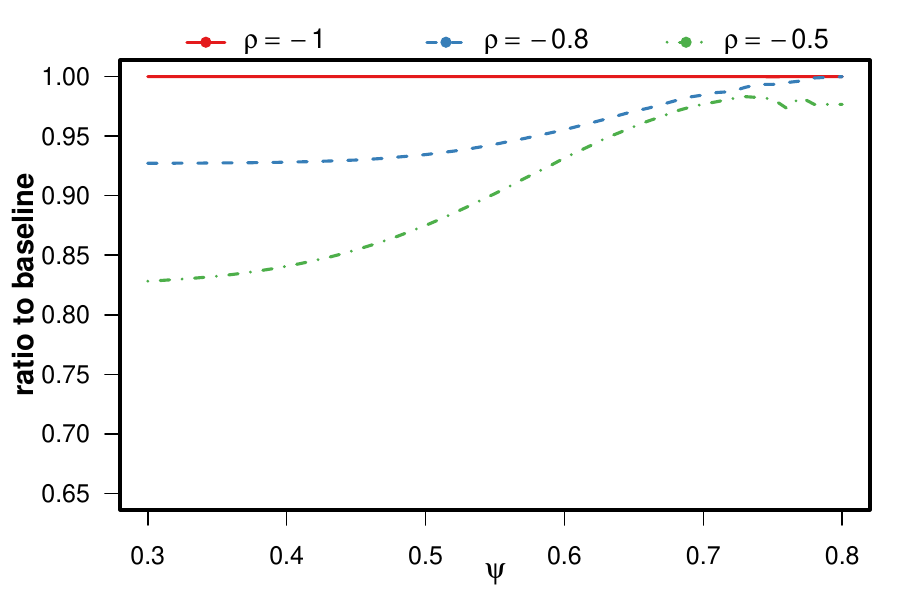}
  \hfill
  \includegraphics[width=0.32\textwidth]{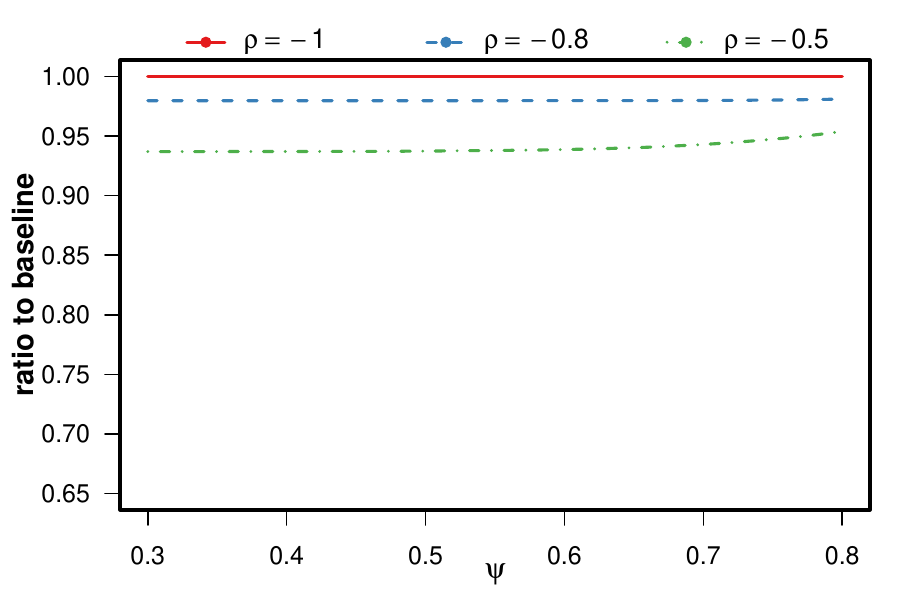}  
  \caption{
  \textit{Upper.} RtB for different candidate score distributions under $\theta \ci s$. 
  Left: good-$k$ solution for fatigue with $\epsilon_1$ over $\psi$. 
  Center: good-$k$ solution for fatigue with $\epsilon_2$ over $\psi$. 
  Right: best-$k$ solution for fatigue with $\epsilon_1$ over $q$.
  \textit{Lower.} RtB for different $\rho$'s (i.e., under $\theta \not\!\perp\!\!\!\perp s$) over $\psi$. 
  Left: good-$k$ solution for fatigue with $\epsilon_1$ and $tN(0.5, 0.02)$ distribution.
  Center: good-$k$ solution for fatigue with $\epsilon_2$ and $tN(0.5, 0.02)$ distribution.
  Right: good-$k$ solution for fatigue with $\epsilon_1$ and $tN(1, 0.05)$ distribution.
  }
  \label{fig:4}
\end{figure*}

\subsubsection{Fatigued scores.}
Beyond the triplet, for the fatigued scores of $h_h$, we fix $\lambda=1$, hence $\Phi(t) = t$, and define:
\begin{itemize}
    \item 
    $\epsilon_1 \sim \mathcal{N}(0, \, (0.005 \cdot (t-1))^2)$, i.e., with constant expectation and standard deviation of $0.005 \cdot (t-1)$.
    \item 
    $\epsilon_2 \sim \mathcal{N}(-0.005 \cdot (t-1), \, (0.001 \cdot (t-1))^2)$, i.e., with decreasing expectation and smaller standard deviation than $\epsilon_1$.
\end{itemize}
The assumptions \textit{A1} and \textit{A2}, and defining fatigue as a linear term in the final score \eqref{eq:BiasedScoresForHh} make the scoring function trivial in this setup. We purposely take for granted the truthful evaluation of $\candidatesset$ to focus on how $\theta$ and $h_h$'s fatigue affect the selected $k$ candidates.

\subsubsection{Evaluation metrics.} 
We consider the solution $S^k_{\besttext}$ of the best-$k$ problem \eqref{eq:fair_objective_all_screener} for $h_a$ (Algorithm~\ref{algo:Examination}) as \textit{the baseline solution} and compare it to the solutions of the good-$k$ problem \eqref{eq:fair_objective_U_psi} under $h_a$ (Algorithm~\ref{algo:Cascade}) 
and of best-$k$ and good-$k$ problems under $h_h$ (resp., Algorithms 3 and 4 in Appendix)
We define two metrics to capture how close is a compared solution to the baseline solution:
\begin{itemize}
    \item 
    \textit{Ratio to baseline} (RtB) is the ratio of $U^k_{\addtext}$ between the compared solution and the baseline solution.
    When calculating the utility of $h_h$, we use the truthful scores, not the fatigued scores, to compare to the baseline.
    For the solution $S_{\goodtext}^k$ under $h_a$, e.g., it is $U^k_{\addtext}(S_{\goodtext}^k) / U^k_{\addtext}(S^k_{\besttext})$. 
    \item 
    \textit{Jaccard similarity} (JdS) is the proportion of candidates in both the compared and baseline solutions over those in at least one of the two solutions. 
    For the solution $S_{\goodtext}^k$ under $h_a$, e.g., it is $|S_{\goodtext}^k \cap S^k_{\besttext}|\, / \, |S_{\goodtext}^k \cup S^k_{\besttext}|$. 
\end{itemize}
The RtB captures whether the compared solution achieves the same utility as the baseline solution as measured by $U^k_{\addtext}$, while the JdS captures the overlap in candidates between the compared solution and the baseline solution. 
For both metrics, the closer the ratio is to $1$, the better the compared solution approximates the baseline solution in terms of, respectively, utility and composition.
The plots report the mean output of these metrics over all the runs.


\subsection{Experiments without Fatigue}
\label{sec:Experiments.Metrics.outFatigue}

Let us start with $h_a$ for clarifying the relation between the best-$k$ (Algorithm~\ref{algo:Examination}) and good-$k$ (Algorithm~\ref{algo:Cascade}) solutions,  
and are interested in whether these differ due to $\theta$ since the best-$k$ requires a full search while the good-$k$ allows for a partial search of $\candidatesset$.
To do so, we look at the impact of different individual scores distributions at the variation of $\psi$.
We consider the case for $\theta \ci s$ and focus on the good-$k$ as $\psi$ is specific to this problem.
Based on Figure~\ref{fig:1} (upper, center and right),
we find that the good-$k$ approximates the best-$k$ solution, especially when there is a low probability of having top candidates in $\candidatesset$, as $\psi$ increases and screening becomes more selective.
In fact, the $k$ first good-enough candidates essentially become the $k$ top candidates as $\psi$ increases.
The symmetric distribution of scores (in red) illustrates clearly this point: having few top candidates forces $h_a$ to explore more $\candidatesset$ following $\theta$, especially under an increasing $\psi$.
The opposite holds for the other two distributions of scores, asymmetric (in blue) and increasing (in green), which are more resilient to $\psi$ as each represents a higher concentration of top candidates in $\candidatesset$: having many top candidates makes it difficult for $h_a$ to select the $k$ top candidates under a partial search.

As the RtB and JdS metrics show in Figure~\ref{fig:1} (upper, center and right), $h_a$ still achieves significant utilities under the asymmetric and increasing score distributions relative to the symmetric score distribution but is unlikely to derive the same $k$ selected candidates under a partial search w.r.t.~a full search of $\candidatesset$ despite a large $\psi$.
Clearly, as noted in Remark~\ref{remark:ISOandPS}, where the $k$ top candidates fall within $\theta$ determines if they are selected or not by $h_a$ under the partial search.
Further, the position bias becomes more prevalent when $\candidatesset$ has many top candidates as even its best candidate may never be selected by $h_a$ if it lies at the bottom of $\theta$.

We also consider the case for $\theta \not\!\perp\!\!\!\perp s$, which further illustrate the role of $\theta$.
We study the impact $\rho$ where $\rho = -1$ essentially implies a ranked $\theta$.
Figure~\ref{fig:1} (lower, center and right) reports the RtB metric for the symmetric and increasing score distributions for different $\rho$'s and increasing $\psi$.
In short, an ISO that negatively correlates with candidate quality greatly reduces the difference in utility between the good-$k$ and best-$k$ solutions.
We find that the good-$k$ solution approximates well the best-$k$ solution already for $\rho=-0.5$, while for $\rho=-1$ the two solutions are the same.
These results are expected as $\theta$ represents the best-$k$ solution depending on $\rho$'s strength.
For instance, under $\rho=-1$, the $k$ first good-enough candidates are also the $k$ best candidates in $\candidatesset$. 

Further, we study the impact of changing the number of $n$ candidates in $\candidatesset$ and $k$ candidates to be selected.
Figure~\ref{fig:1} (lower, left) show the RtB for the symmetric score distribution under $\theta \ci s$, though the result hold for the other two distribution as well as under $\theta \not\!\perp\!\!\!\perp s$. 
We compare $n=120, k=6$ to $n=400, k=20$ and $n=30, k=6$ (with ratio of selected $k/n$, resp., of $0.05$ and $0.2$). 
The plots show that changes in the ratio $k/n$ affect the metric, in particular, a larger ratio leads good-$k$ to better approximate best-$k$ for a same $\psi$. 
Clearly, the influence of $\theta$ diminishes as $k/n$ increases.
Furthermore, we study the impact of changing $q$.
Here, results are expected given our underlying \textit{A1} and \textit{A2} assumptions, finding that under $\theta \ci s$, $q$ does not affect the relative strengths of best-$k$ and good-$k$ solutions. 
See Appendix.

\subsection{Experiments with Fatigue}
\label{sec:Experiments.Metrics.withFatigue}

Let us now focus on $h_h$, and start by studying whether the screener's fatigue impacts the utility relative to the baseline solution (namely, Algorithm~\ref{algo:Examination}). 
We compare the best-$k$ and good-$k$ solutions with fatigue (Algorithms 3 and 4 in Appendix)
to such baseline.
We consider the case of $\theta \ci s$.
Figure~\ref{fig:4} (upper, left) shows the RtB metric for the three score distributions for the good-$k$ solution with fatigued scores due to $\epsilon_1$. 
Notice that in Figure~\ref{fig:1} (upper, center), for the asymmetric (in blue) and increasing (in green) score distributions, there is no considerable difference w.r.t. the case without fatigue.
In Figure~\ref{fig:4} (upper, left), instead, for the symmetric score distribution (in red) there is a considerable decrease under high $\psi$ values, which can be attributed to a low number of top scores on which $\epsilon_1$'s has a large effect.
For the other two score distributions, there are enough top scores such that $\epsilon_1$ does not change the top score distribution.
Since the RtB metric captures achieving the utility of the baseline model, $h_h$ is still able to reach high utility solutions under a partial search when $\candidatesset$ has many top candidates.
Note, thought, that as $\psi$ increases and screening becomes more selective, $h_h$ becomes more tired under $\theta$ as it needs to further search $\candidatesset$ to achieve $k$. 
Naturally, having enough top candidates reduces the need for $h_h$ to further search $\candidatesset$ as $\psi$ increases. 
Similar to Figure~\ref{fig:4} (upper, left), Figure~\ref{fig:4} (upper, center) considers the fatigued scores due to $\epsilon_2$. 
The effect on the symmetric distribution (in red) is not present here due to the lower standard deviation of $\epsilon_2$. 
The bias of $\epsilon_2$ does not impact screener utility too much under a $\candidatesset$ with low candidate quality.
Overall, under $\theta \ci s$, variance appears more relevant than bias in the case of low probability for top scores.
This result illustrates the importance of how we define fatigue.

Further, Figure~\ref{fig:4} (upper, right) shows the RtB for the best-$k$ solution at the variation of the quota $q$ that constrains the set selection based on group-level fairness goals. 
Here, there is a considerable and constant loss in screener utility under fatigue, which is more consistent for the symmetric score distribution (in red). 
The RtB is lower than in the case of the good-$k$ with fatigue for $\psi \geq 0.5$. 
This result means that, for the symmetric distribution, the good-$k$ solution with fatigue has better utility than the best-$k$ solution with fatigue. 
Such comparison shows the impact on $h_h$ from following for too long $\theta$, and indicate that one search procedure might be preferred over the other one in certain settings involving $h_h$.

We now consider the case for $\theta \not\!\perp\!\!\!\perp s$, looking at the impact of $\rho$ on $\theta$ with again a focus on the good-$k$ solution.  
Figure~\ref{fig:4} (lower, left) considers the symmetric score distribution (in red) where, for the lower half of $\psi$'s, the lines are similar to the analogous case without fatigue shown in Figure~\ref{fig:1} (lower, center). 
For the higher half of $\psi$'s, instead, there is a decrease in the RtB. 
Again, these results are due to the low probability of top scores for which the effects of the bias due to $\epsilon_1$ is not counter-balanced by $\rho$.
Such an effect does not appear for $\epsilon_2$ nor for $\epsilon_1$ under the increasing score distribution. 
In fact, Figure~\ref{fig:4} (lower, center) and (lower, right) closely resemble those in Figure~\ref{fig:1} (lower, center) and (lower, right), respectively.
It means that fatigue does not have an impact on the utility of the good-$k$ solution if there are sufficiently many top scores or a sufficiently small variability of the fatigue.
This last result points at the importance of providing a $\theta$ to the human screener with some information about candidate quality. 
Intuitively, under a partial search procedure and the threat of position bias through $\theta$, we would like to decrease $h_h$'s fatigue by minimizing its need to further search $\candidatesset$, which reinforces the role of IAS in the ISO problem.

%
%

\section{Conclusion}
\label{sec:Discussion}

In this work, we presented the \textit{initial screening order} (ISO) as a parameter of interest; 
defined two formulations under distinct utility models of the set selection problem, the best-$k$ and the good-$k$, with their corresponding algorithmic implementations; 
and introduced a human-like screener to study the effects of the ISO on a human user. 
We also provided a simulation framework flexible enough to study and model multiple screening scenarios.
Our analysis confirms the fairness and optimality impact of the ISO, motivated by the risk of position bias, on the set of $k$ selected candidates by an algorithmic or human-like screener via an IAS. 

The simulations show a complex relations between the best-$k$ and good-$k$ problems.
Our results are limited by the functional assumptions made for formulating the two problems and screeners, in particular, the human-like screener.
Future work should explore alternative utility models and fatigue terms, while still relying on the current experimental framework.
An alternative formulation to fatigue, e.g., could involve a human-like screener that rests while searching over the ISO. 
We see recurrent survival models \cite{DBLP:conf/www/ChandarTMPSWCLJ22} well suited for this task.
Future work should also explore theories on human decision-making (e.g., \cite{DBLP:books/daglib/0033056,DBLP:conf/chi/CarabanKGC19, DBLP:journals/isr/AdomaviciusBCZ13}), or the use of the simulations framework for testing for optimal parameters (e.g., deriving a minimum score $\psi$ for which best-$k$ and good-$k$ problems coincide). 
That said, given that it is costly and time-consuming to run real candidate screening experiments, especially at the same scale of Section~\ref{sec:Experiments},
we view our work as another example \cite{DBLP:conf/fat/IonescuHJ21, DBLP:journals/corr/abs-2006-09663, DBLP:conf/fat/BountouridisHMM19, Bokanyi2020_Understanding, Schelling1971_Dynamic} of how simulations can be useful tools for studying the fairness and optimality of real-world decision-making processes involving ML-based systems \cite{DBLP:journals/air/MosqueiraReyHABF23}.

We conclude with two takeaways from our analysis for user search behavior. 
First, defining the proper problem formulation is important for understanding the impact of the ISO on the selected candidates, which reflects on the search procedures of the screeners regardless of their kind.
Second, once the search procedure is clear, it is important to understand how the screener behaves as it searches over the ISO.
These takeaways have direct implications for practitioners.
They might seem obvious ex-post, though they are supported by an extensive analysis that accounts for several factors that only become clear under modeling and experiments.

%
%

\section*{Acknowledgments} 
Work supported by the European Union (EU)’s Horizon Europe research and innovation program for the project FINDHR (g.a. 101070212), and under the Horizon 2020 Marie Sklodowska-Curie Actions research and innovation program  for the project NoBIAS (g.a. 860630). Views and opinions expressed are those of the authors only and do not necessarily reflect those of the EU. Neither the EU nor the granting authority can be held responsible for them.

\clearpage

\section*{Ethical Considerations}
We did not face ethical challenges when drafting the paper. Results are based on simulated data intended to illustrate our theoretical analysis. During our collaboration with company G, in particular, which occurred before the drafting of this paper, we followed G's ethical guidelines at all times. We concluded our collaboration with G with an internal report that we presented and discussed with all stakeholders. No sensitive data (or data at all) from company G was used for this work. At no point did we receive monetary compensation from G. The views reflected are entirely our own. Further, we believe that this work shows the importance of considering the human user in the formulation of the candidate screening problem. We stress that our distinction between an algorithmic screener and a human-like screener is to show the importance of considering the latter kind and not to endorse the former kind. We strongly believe that candidate screening is a complex, human-dependent and human-centered process that should not be left as an automated decision-making problem.

\bibliographystyle{ACM-Reference-Format}
\balance
\bibliography{references}

\clearpage

\appendix
\label{Appendix}

\begin{figure*}[!t]
\scalebox{0.95}{
\begin{minipage}[t]{0.45\textwidth}
\begin{algorithm}[H]
\caption{HumanLikeExaminationSearch}
\label{algo:HumanExamination}
\begin{algorithmic}[1]
\Require $n$, $\theta$, $k$, $q$, $\Phi$
\Ensure $S^k_\besttext$
\State $q^* \, \gets$ \texttt{round(}$q \cdot k$\texttt{)}; $r^* \, \gets k - q^*$
\State $\mathit{scores} \, \gets [ s(\mathbf{X}_{\theta(t)})+\epsilon(\Phi(t-1))\ \mathit{for}\ t = 1, \ldots, n ]$
\State $\tau \gets \texttt{argsortdesc(} \mathit{scores}$\texttt{)}
\State $i \, \gets 1$; $k^* \, \gets 0$; $Q \, \gets \{ \}$; $R \, \gets \{ \}$
\While{$k^* < k$} 
    \State $c \; \gets \theta(\tau(i))$; $i \, \gets i + 1$
    \If{$W_c == 0$ \textbf{and} \texttt{len(}$R$\texttt{)}$== r^*$ }
       \State \textbf{continue}
    \EndIf
    \State $k^* \, \gets k^* + 1$
        \If{$W_c==1$ \textbf{and} $\texttt{len(}Q\texttt{)} < q^*$}
             \State $Q \, \gets Q \cup \{ c \}$
        \Else
             \State $R \, \gets R \cup \{ c \}$
       \EndIf
\EndWhile
\Return $Q \cup R$
\end{algorithmic}
\end{algorithm}
\end{minipage}
}
\hspace{0.05\textwidth}
\scalebox{0.95}{
\begin{minipage}[t]{0.425\textwidth}
\begin{algorithm}[H]
\caption{HumanLikeCascadeSearch}
\label{algo:HumanCascade}
\begin{algorithmic}[1]
\Require $n$, $\theta$, $k$, $q$, $\psi$, $\Phi$
\Ensure $S_{\goodtext}^k$
\State $q^* \, \gets$ \texttt{round(}$q \cdot k$\texttt{)}; $r^* \, \gets k - q^*$
\State $i \, \gets 1$; $k^* \, \gets 0$; $Q \, \gets \{ \}$; $R \, \gets \{ \}$; $t \, \gets 1$
\While{$k^* < k$} 
    \State $c \; \gets \theta(i)$; $i \, \gets i + 1$
    \If{$W_c == 0$ \textbf{and} \texttt{len(}$R$\texttt{)}$== r^*$ }
       \State \textbf{continue}
    \EndIf
    \State $Y_c \; \gets s(\mathbf{X}_c) + \epsilon(\Phi(t-1)$); $t \, \gets t + 1$
    \If{$Y_c \geq \psi$}
      \State $k^* \, \gets k^* + 1$
        \If{$W_c==1$ \textbf{and} $\texttt{len(}Q\texttt{)} < q^*$}
             \State $Q \, \gets Q \cup \{ c \}$
        \Else
             \State $R \, \gets R \cup \{ c \}$
       \EndIf
      \EndIf
\EndWhile
\Return $Q \cup R$
\end{algorithmic}
\end{algorithm}
\end{minipage}
}
\caption{The algorithmic implementations for the best-$k$ and good-$k$ problems revisited under the human-like screener $h_h$.}
\end{figure*}
\begin{figure*}[t]
  \centering
  \includegraphics[width=0.32\textwidth]{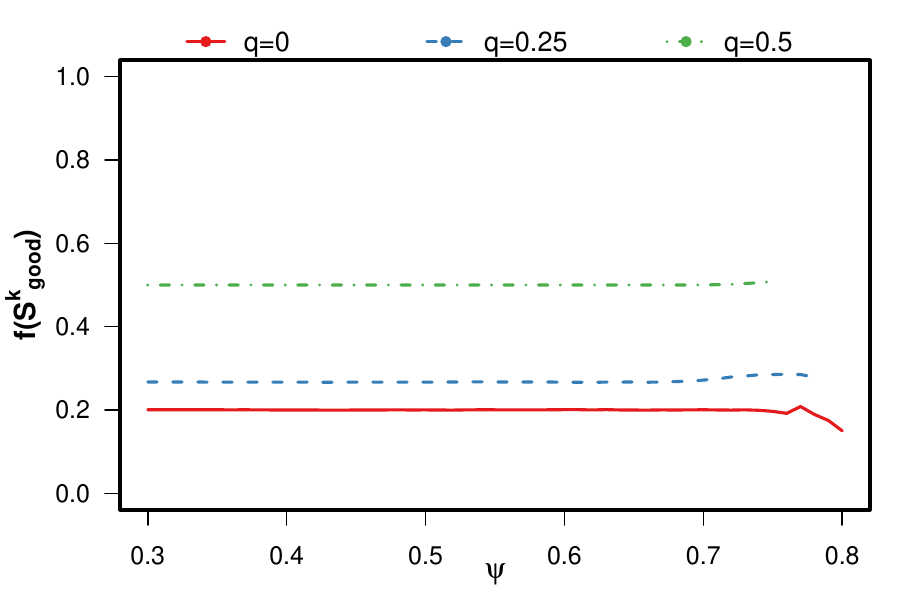}
  \hfill
  \includegraphics[width=0.32\textwidth]{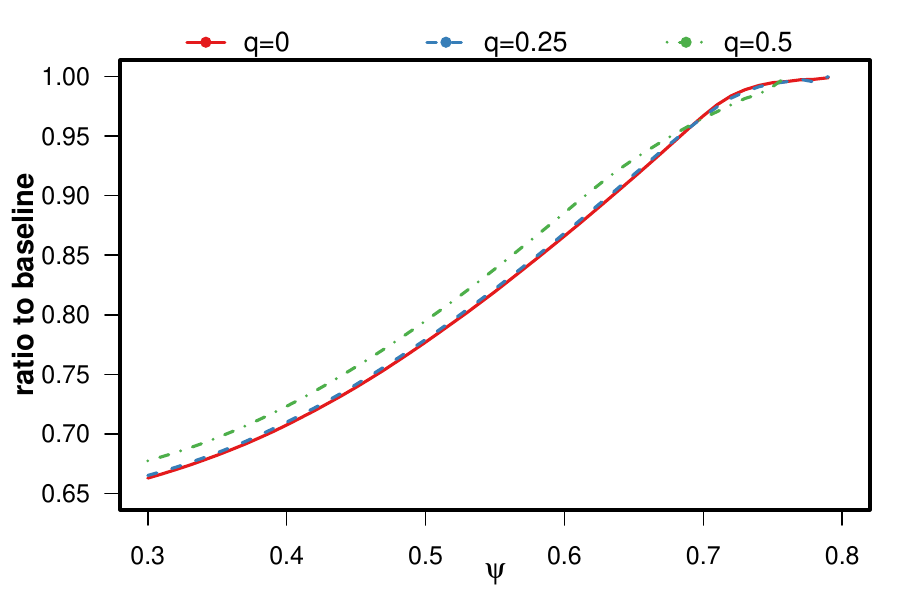}
  \hfill
  \includegraphics[width=0.32\textwidth]{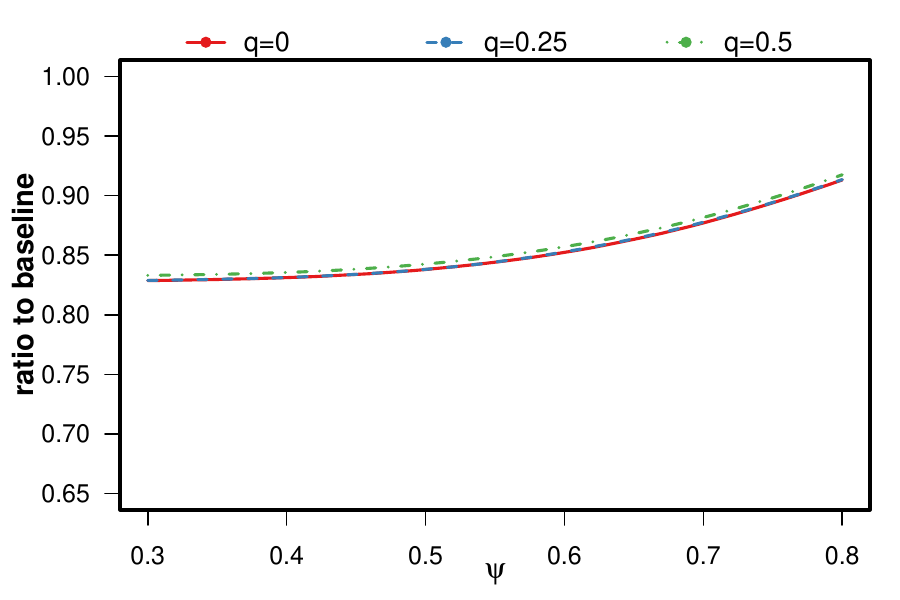} 
  \caption{
  Left: fraction of protected candidates in the solution of good-$k$ for different representational quotas $q$, for the $tN(0.5, 0.02)$ score distribution and with setting $n=400$, $k=20$, $\theta \ci s$. 
  Center: RtB for different representational quotas $q$, for the $tN(0.5, 0.02)$ score distribution and with setting $n=400$, $k=20$, $\theta \ci s$. 
  Right: RtB for different representational quotas $q$, for the $tN(1, 0.05)$ score distribution and with setting $n=400$, $k=20$, $\theta \ci s$.
  }
\label{fig:3}
\end{figure*}

\section{Collaborating with G}
\label{Appendix.MoreOnG}

Candidate screening at G represented both a time-consuming, repetitive task prone to human error and a sensitive, high-risk task requiring human oversight.
Therefore, the option of full automation was not possible. 
The focus was, thus, on understanding and modeling the search of the HR officer via the hiring platform.
Importantly, under the realistic risk of position bias affecting candidate evaluations via the hiring platform, we wanted to study the influence of the \textit{initial screening order} (ISO) on the set of suitable candidates chosen by the HR officer.

\paragraph{Overall experience.}
The collaboration lasted for four months.
Due to the COVID-19 pandemic, it was a hybrid collaboration.
During this time, we mostly interviewed the HR officers to understand their tasks, constraints, and methodologies, often shadowing them during screening sessions.
We were also granted access to Taleo by Oracle, the platform used by HR for managing the hiring pipeline.
This allowed us to experience for ourselves the patterns we observed among the HR officers. These patterns resulted in the five stylized facts in Section~\ref{sec:Generali}. 

We interacted, in particular, with five HR officers specialized in screening applications for technical roles within G, such as the roles of data scientist and front-end developer.
These HR officers had to process considerable amounts of information within a time constraint. Based on what we observed, most candidate pools (except for very senior profiles like, e.g., director of data science) consisted of hundred of applications. 
These HR officers were involved in screening multiple candidate pools with similar deadlines within the same week.
It became apparent to us, specially for candidate screening, how time-consuming and human-dependent was the hiring process at G. 

We discussed our observations, often on a bi-weekly basis, to members of both the HR and Advanced Analytics (AA) teams.
We emphasize that we simply collaborated with these teams as equals, sharing the goal of understanding G's hiring process and whether it was suitable or not for (fair) automation. 
We specifically use the wording “stylized facts” in Section~\ref{sec:Generali} to emphasize that we draw inspiration from the collaboration with G in formulating the ISO problem rather than ``hard facts'' about G derived from an observational study. In fact, the first draft of this paper occurred about one year after our collaboration with G had concluded.

\paragraph{Deliverables.}
We concluded the collaboration with a report to AA that formalized G's candidate screening process as a ranking problem, evaluated the potential fairness implications, and assessed the risk and benefits of automation.
The report was discussed with a wider audience within G in a series of presentations hosted by the AA and HR teams. 
The report focused mainly on candidate screening.
It is worth mentioning also that throughout the collaboration, we followed G's strict ethical guidelines at all times.
No sensitive data (or data at all) from G was used for this work.
This paper is not a deliverable specific to G.

\paragraph{The hiring pipeline.}
Hiring at G consists mainly of three phases. With respect to the stylized facts, these are concerned with the candidate screening phase or phase two.
These three phases are the following:
\begin{itemize}
    \item In \textit{phase one}, the HR builds a candidate pool for the job opening. Candidates submit their CVs, complete a multiple-choice questioner, and write a motivation letter. Sensitive information, such as gender, ethnicity, and age, is also provided or it can be inferred. The candidate pool is stored in a database platform.
    \item In \textit{phase two}, the HR officer reduces the candidate pool into a smaller pool of suitable candidates. The HR officer determines candidate suitability based on each candidate's profile using a set of minimum basic requirements.
    \item In \textit{phase three}, the chosen candidates are interviewed by HR and the team offering the job. 
    It is common for the hiring team to also prepare a use case for the candidates.
    The best candidates receive an offer. If no candidates are hired, HR resorts to the runner-up candidates from phase two and repeat phase three.
\end{itemize}
Note that the above represent G's hiring pipeline during the collaboration. We do not know if this hiring pipeline is still the case today, though it is not important for the purpose of this paper.

\section{Supplementary Material}
\label{Appendix.SupplementaryMaterial}

\subsection{Additional Discussion on the Utility Model in Section~\ref{sec:good-k}}
\label{Appendix.NaiveUtilityGoodk}

First, we present Example~\ref{ex:diff_fractions_prot} below. 
It shows that for the utility model \eqref{eq:AlternativeUtility} for the good-$k$ problem, in the general case, i.e.,~$f(S^k) \geq q$, there can be two solutions, but with different fractions of the protected group.

\begin{example}
\label{ex:diff_fractions_prot}
    Consider $n=3, k=2, q=0.5$. Assume three eligible candidates and $\theta(1) = c_1, \theta(2) = c_2, \theta(3) = c_3$ with $W_1 = 0$ and $W_2 = W_3 = 1$. Both $S' = \{c_1, c_2\}$ and $S'' = \{c_2, c_3\}$ are solutions of (\ref{eq:fair_objective_U_psi}) with $U^k_{\psi} \big(S', \theta\big) = U^k_{\psi} \big(S'', \theta\big) = 2$. However, $f(S') = 0.5$ and $f(S'') = 1$. Intuitively, $S'$ is obtained by strictly iterating over $\theta$.
\end{example}

\noindent
We now motivate the utility model \eqref{eq:AlternativeUtility} used in the good-$k$ problem.
As a simple alternative utility function in the good-$k$ setting consider the following utility model:
\begin{equation}
\label{eq:AlternativeUtility2}
    \hat{U}^k_{\psi}\big( S^k, \theta \big) = \left\{
    \begin{array}{ll}
        n - \max_{c \in S^k} \theta^{-1}(c) & \text{if} \  \forall c \in S^k \  s(\mathbf{X}_c) \geq \psi   \\
        0 & \text{otherwise.}
    \end{array} \right.
\end{equation}
where we use $\hat{U}_\psi^k$ (i.e., the $U$ hat) to differentiate from the utility model $U_\psi^k$ considered in \eqref{eq:AlternativeUtility}.

Intuitively, in the above utility definition the screener wants to find as quickly as possible a set of $k$ eligible candidates.
Therefore, if $S^k$ contains only eligible candidates, the utility of $h$ selecting $S^k$ under $\theta$ is expressed by the number of candidates past the last one who was screened, i.e.,~the ``saved effort" of the screener $h$.
Despite the simplicity of \eqref{eq:AlternativeUtility2}, the above utility model \eqref{eq:AlternativeUtility2} is not suitable to properly account for the intended good-$k$ problem.
To observe this last point, consider Example~\ref{ex:AltU2}.

\begin{example}
\label{ex:AltU2}    
    Let $n=3, k=2, q=0.5$. Assume three eligible candidates and $\theta(1) = c_1, \theta(2) = c_2, \theta(3) = c_3$ with $W_1 = W_2 = 0$ and $W_3 = 1$. It turns out that both $S' = \{c_1, c_3\}$ and $S'' = \{c_2, c_3\}$ maximize the utility \eqref{eq:AlternativeUtility2} and satisfy the fairness constraint $q$.
\end{example}

Following up on Example~\ref{ex:AltU2}, why should have been $c_2$ considered, and then returned in $S''$, if $c_1$ already meets the minimum basic requirement? 
A reason for doing that is a variant of our good-$k$ problem in which the screener $h$ keeps evaluating non-protected candidates in $\mathcal{C}$, even if their quota is reached but the one of protected candidates is not yet reached, for the purpose of keeping the best ones found so far. 
We do not consider such a variant in this paper.
For this reason, we introduce the penalty function \eqref{eq:Penalty} in \eqref{eq:AlternativeUtility} presented in Section~\ref{sec:good-k}.

\subsection{The Two Search Procedures under a Human-Like Screener}
\label{Appendix.HumanAlgorithms}

We update the \textit{ExaminationSearch} and \textit{CascadeSearch} and their corresponding Algorithm \ref{algo:Examination} and Algorithm \ref{algo:Cascade} from Section~\ref{sec:ProblemFormulation.Algorithms} under the human-like screener $h_h$ from Section~\ref{sec:HumanScreener}. 
We incorporate the \textit{fatigued scores} formulation from Section~\ref{sec:HumanScreener.BiasedScores} into both algorithms, resulting in a human-like \textit{HuamnExaminationSearch} (Algorithm~\ref{algo:HumanExamination}) and a human-like \textit{HumanCascadeSearch} (Algorithm~\ref{algo:HumanCascade}).
In comparison to the algorithmic screener $h_a$, the main difference here is that both algorithms compute the fatigued score for candidate $c \in \mathcal{C}$:
\begin{equation}
    Y_c = s(\mathbf{X}_c) + \epsilon
\end{equation}
where $\epsilon$ is a random variable depending on the accumulated fatigue $\Phi$ of $h_h$.
In both algorithms \ref{algo:HumanExamination} and \ref{algo:HumanCascade}, by requiring the fatigue component $\Phi$, we also require a specific modeling choice for $\epsilon$ which is a probabilistic function of the accumulated fatigue $\Phi$. As discussed in Section~\ref{sec:HumanScreener.BiasedScores}, $\epsilon$ can be modeled as either $\epsilon_1$ or $\epsilon_2$.

We stress once again that other formulations for $\epsilon$ are possible.
These formulations are compatible with Algorithms~\ref{algo:HumanExamination} and \ref{algo:HumanCascade} as long as $\epsilon$ is treated as a random variable that is drawn each time $h_h$ evaluates candidate $c$.
These formulations can be implemented as with $\epsilon_1$ and $\epsilon_2$, meaning by providing the corresponding probability distribution for the intended accumulated fatigue $\Phi$.

\section{Additional Experiments}
\label{Appendix.MoreExperiments}

In this section, we present additional figures and corresponding discussions relating to the experimental analysis of the algorithmic screener $h_a$ from Section~\ref{sec:Experiments.Metrics.outFatigue}.

Let us consider the quota parameter $q$, thus far set to $q=0.5$ over a population with a fraction of protected candidates set to $\mathit{pr}=0.2$. 
Since we assumed that $W$ is independent from both scores and the ISO, the fraction of protected group in the solutions of best-$k$ and good-$k$ is, on average, $\mathit{min}\{q, \mathit{pr}\}$. 
Figure~\ref{fig:3} (left) shows this result in the solution for good-$k$. 
A less trivial question is whether $q$ is also not affecting the evaluation metrics: e.g., whether the quota $q$ changes the ratio to baseline? 
Figure~\ref{fig:3} (center, right) show that this is not the case in two experimental settings. 
Again, this result is theoretically implied by the independence of $W$ with scores and initial order. 
In summary, under $\theta \ci s$, the quota $q$ in the best-$k$ and good-$k$ problem does not affect the relative strengths of their solutions.

%
%

\end{document}